\documentclass[sigconf]{acmart}
\usepackage{pifont}
\usepackage{float}
\usepackage{booktabs}
\usepackage{multirow}
\usepackage{longtable}
\usepackage{hyperref} 
\usepackage{url}
\usepackage{color,colortbl}
\usepackage{balance}
\definecolor{lightred}{RGB}{251,49,153}

\AtBeginDocument{%
  \providecommand\BibTeX{{%
    \normalfont B\kern-0.5em{\scshape i\kern-0.25em b}\kern-0.8em\TeX}}}
\copyrightyear{2025}
\acmYear{2025}
\setcopyright{acmlicensed}
\acmConference[MM '25] {Proceedings of the 33rd ACM International Conference on Multimedia}{October 27--31, 2025}{Dublin, Ireland.}
\acmBooktitle{Proceedings of the 33rd ACM International Conference on Multimedia (MM '25), October 27--31, 2025, Dublin, Ireland}
\acmISBN{979-8-4007-2035-2/2025/10}
\acmDOI{10.1145/3746027.3754951}
\begin{document}

\title{Beyond Artificial Misalignment: Detecting and Grounding Semantic-Coordinated Multimodal Manipulations}

\author{Jinjie Shen}
\orcid{0009-0000-0334-1669} 
\affiliation{%
  \institution{Hefei University of Technology}
  \city{Hefei}
  \state{Anhui}
  \country{China}
}
\email{shenjinjie22@gmail.com}

\author{Yaxiong Wang$^{*}$}
\thanks{$^{*}$Corresponding authors.}
\orcid{0000-0001-6596-8117}
\affiliation{%
  \institution{Hefei University of Technology}
  \city{Hefei}
  \state{Anhui}
  \country{China}
}
\email{wangyx@hfut.edu.cn}

\author{Lechao Cheng}
\orcid{0000-0002-7546-9052}
\affiliation{%
  \institution{Hefei University of Technology}
  \city{Hefei}
  \state{Anhui}
  \country{China}
}
\email{chenglc@hfut.edu.cn}

\author{Nan Pu}
\orcid{0000-0002-2179-8301}
\affiliation{%
  \institution{University of Trento}
  \city{Trento}
  \state{Trentino}
  \country{Italy}
}
\email{nan.pu@unitn.it}

\author{Zhun Zhong$^{*}$}
\orcid{0000-0002-8202-0544}
\affiliation{%
  \institution{Hefei University of Technology}
  \city{Hefei}
  \state{Anhui}
  \country{China}
}
\email{zhunzhong007@gmail.com}

\renewcommand{\shortauthors}{J. Shen, Y. Wang, et al.}
\begin{abstract}
 The detection and grounding of manipulated content in multimodal data has emerged as a critical challenge in media forensics. While existing benchmarks demonstrate technical progress, they suffer from misalignment artifacts that poorly reflect real-world manipulation patterns: practical attacks typically maintain semantic consistency across modalities, whereas current datasets artificially disrupt cross-modal alignment, creating easily detectable anomalies. To bridge this gap, we pioneer the detection of semantically-coordinated manipulations where visual edits are systematically paired with semantically consistent textual descriptions. Our approach begins with constructing the first Semantic-Aligned Multimodal Manipulation (SAMM) dataset, generated through a two-stage pipeline: 1) applying state-of-the-art image manipulations, followed by 2)  generation of contextually-plausible textual narratives that reinforce the visual deception. Building on this foundation, we propose a Retrieval-Augmented Manipulation Detection and Grounding (RamDG) framework. RamDG commences by harnessing external knowledge repositories to retrieve contextual evidence, which serves as the auxiliary texts and encoded together with the inputs through our image forgery grounding and deep manipulation detection modules to trace all manipulations. Extensive experiments demonstrate our framework significantly outperforms existing methods, achieving 2.06\% higher detection accuracy on SAMM compared to state-of-the-art approaches. The dataset and code are publicly available at \textcolor{lightred}{\url{https://github.com/shen8424/SAMM-RamDG-CAP}}
\end{abstract}




\begin{teaserfigure}
    \centering
    \includegraphics[width=0.9\textwidth]{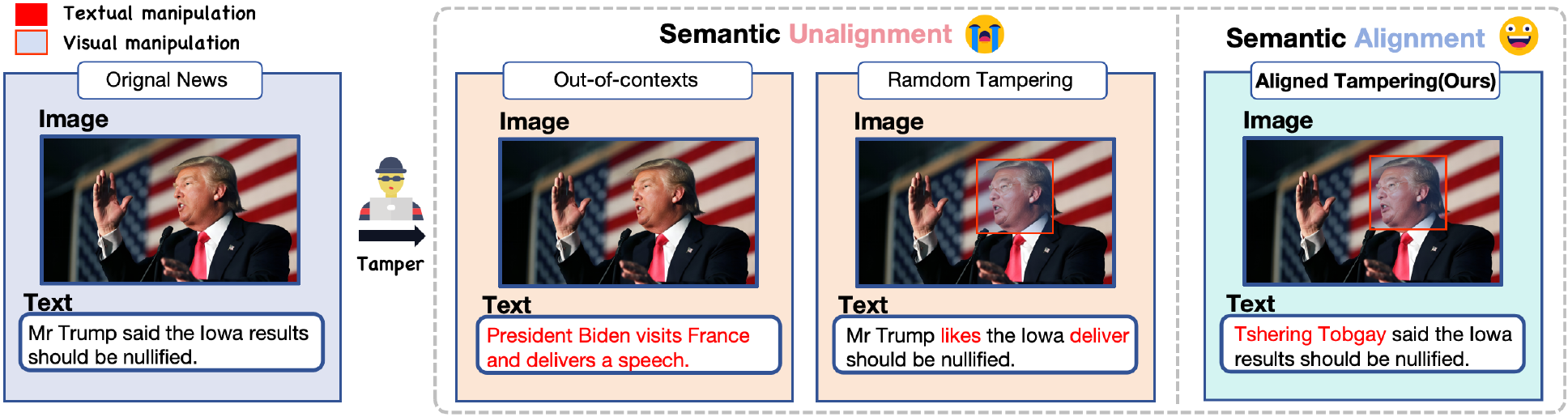}
    \caption{Unlike existing multimodal deepfake datasets where cross-modal semantic alignment is lacking, SAMM proposes semantically aligned fake news, which better reflects real-world scenarios.}
    \label{fig:teaser}
\end{teaserfigure}
\maketitle
\section{Introduction}
The rapid development of generative models has driven significant progress across domains~\cite{intro04,intro05,intro06,entityCLIP,cala}. Concurrently, this technological advancement precipitates critical societal risks, particularly through the synthesis of highly plausible yet falsified media content~\cite{intro07,intro08,intro09,intro10}. Such fabricated information not only erodes public trust through fabricated misleading contents but also engenders systemic vulnerabilities in digital information ecosystems~\cite{deepfake1,deepfake2,deepfake3}.

Many efforts have been made to recognize the fake news in social media~\cite{intro01,intro02,intro03,asap}. In recent years, different types of scenarios of manipulated multimodal news have been studied~\cite{dgm4,newsclipings,miragenews,intro11}.  NewsCLIPings~\cite{newsclipings} uses randomly selected news texts to form out-of-context image-text Pairs.  DGM$^{4}$~\cite{dgm4} employs a random modification strategy as well, which in the visual modality involve using randomly selected non-celebrity faces as substitutes, and in the textual modality involve randomly replacing certain words or segments. Despite these pioneering attempts,  the focused scenarios of these works all have a severe artifact of semantic misalignment. For example, as shown in the top row of Figure~\ref{fig:teaser}: ``\emph{an image of Mr Trump giving a speech is paired with a caption that reads president Biden performing at a concert.}'' This misalignment, on the one hand, renders the fake news detection too easy to cheat the people. On the other hand, the semantic-misaligned news fails to stimulate the practical situations, since the attackers usually maintain the consistency across modalities to  deceive the public. For example, if Biden's face is swapped with Trump's in the visual modality, the corresponding text modality would also reflect Biden being replaced by Trump, as shown in the bottom of Figure~\ref{fig:teaser}.

In response to the aforementioned challenges, we focus on a more practical problem in this work: multimodal manipulation detection and grounding with semantic-alignmend manipulations. To facilitate this research, we present the SAMM (Semantic-Aligned Multimodal Manipulation) - a comprehensive dataset containing 260,970 carefully crafted semantic-coordinated samples.  First, we perform visual alterations through either face swapping~\cite{simswap,infoswap} or facial attribute editing ~\cite{hfgi,styleclip} on celebrity images, considering their heightened social impact and misinformation risks in public domains~\cite{dgm4}. Subsequently, we generate semantically-aligned fake text descriptions that maintain logical consistency with the manipulated visual content. This two-stage manipulation pipeline ensures sophisticated alignment between visual tampering and textual fabrication, creating convincing multimodal forgeries that pose significant detection challenges. 

Semantic-coordinated manipulations reflect real-world cases but pose greater detection challenges, as prior methods fail in such scenarios.
%Previous studies on semantic misalignment detect cross-modal inconsistencies \cite{dgm4,clip,albef}, but these approachs are ineffective in semantically coordinated scenarios.
Notably, human usually check the confused information by conducting the cross-verification with external knowledge, such as using the fact that "Messi is a great football player" to identify the fake news claiming "Messi won the Nobel Prize in Literature." Inspired by this, we propose the \textbf{R}etrieval-\textbf{a}ugmented \textbf{m}anipulation \textbf{D}etection and \textbf{G}rounding (\textbf{RamDG}) framework. First, by integrating a large-scale external knowledge base \textbf{C}eleb \textbf{A}ttributes \textbf{P}ortfolio(\textbf{CAP}) containing information on celebrities from various domains, we design the \textbf{C}elebrity-\textbf{N}ews \textbf{C}ontrastive \textbf{L}earning (\textbf{CNCL}) mechanism to facilitate RamDG in leveraging external knowledge for semantic-level fake news detection akin to human capabilities. Furthermore, we introduce the \textbf{F}ine-grained \textbf{V}isual \textbf{R}efinement \textbf{M}echanism (\textbf{FVRM}) module to enhance the model's ability to accurately localize visually manipulated regions.
 
Our main contributions include: 
\begin{itemize}
    \item Introducing SAMM, a more realistic deepfake dataset featuring multi-modal semantic alignment and purposeful tampering, with a large scale and rich fine-grained annotations to meet training or evaluation needs; 
    \item Proposing CAP, an external knowledge base containing multi-domain celebrity information, enabling logical fake news detection through simple "string matching" integration into existing datasets; 
    \item Presenting RamDG, which outperforms in binary classification and excels in fine-grained tampering localization compared to all current models.
\end{itemize}
%--------------------------------------------------
\section{Related work}
\noindent\textbf{DeepFake Detection.}
Historically, deepfake detection has primarily focused on single modalities, such as text~\cite{relat01,relat02,relat03} or visual~\cite{relat04,relat05,relat06}. 
Within visual modalities, methods are categorized into those based on the spatial domain~\cite{relat07,relat08,relat09} and those based on the frequency domain~\cite{relat10,relat11,relat12}. 
With the advancement of multimodal techniques\cite{vilt,clip,llama2,qwen2-vl,pfan,pfan++,cala}, news content is increasingly presented in multimodal formats, leading to the emergence of recent multimodal detection approaches, including methods based on modality fusion~\cite{vilt,dgm4,hammer++} and those leveraging Vision Language Large Model~\cite{sniffer,fka-owl,miragenews}. Modality fusion methods struggle with small datasets due to limited external knowledge, whereas large language models, despite their rich internal knowledge, face challenges in fine-grained tampering localization. To address these limitations, we propose a hybrid approach that integrates CAP-derived external knowledge for detection and enhances fine-grained localization accuracy.

\noindent\textbf{DeepFake Datasets.}
Existing DeepFake datasets primarily consist of single-modality data focusing on either visual~\cite{relat15,relat16} or textual content~\cite{relat13,relat14}. While some multimodal datasets exist~\cite{miragenews,newsclipings}, they typically adopt either contextually irrelevant pairings (out-of-context pairs)~\cite{newsclipings} or rely entirely on synthetic data produced by generative models~\cite{miragenews}. DGM$^{4}$~\cite{dgm4} addresses these issues to some extent by providing over 230k samples with fine-grained annotations, constructed through modifications of authentic news articles. However, All the datasets mentioned above exhibit two critical limitations: 1) Cross-modal semantic inconsistency; 2) Purposeless tampering. To address these limitations, we propose the \textbf{SAMM Dataset}, a large-scale multimodal dataset with comprehensive fine-grained annotations that better aligns with real-world fake news distribution patterns. 
%------------------------------------------------------------------------
\begin{figure*}[t]
    \centering
    \includegraphics[width=\textwidth]{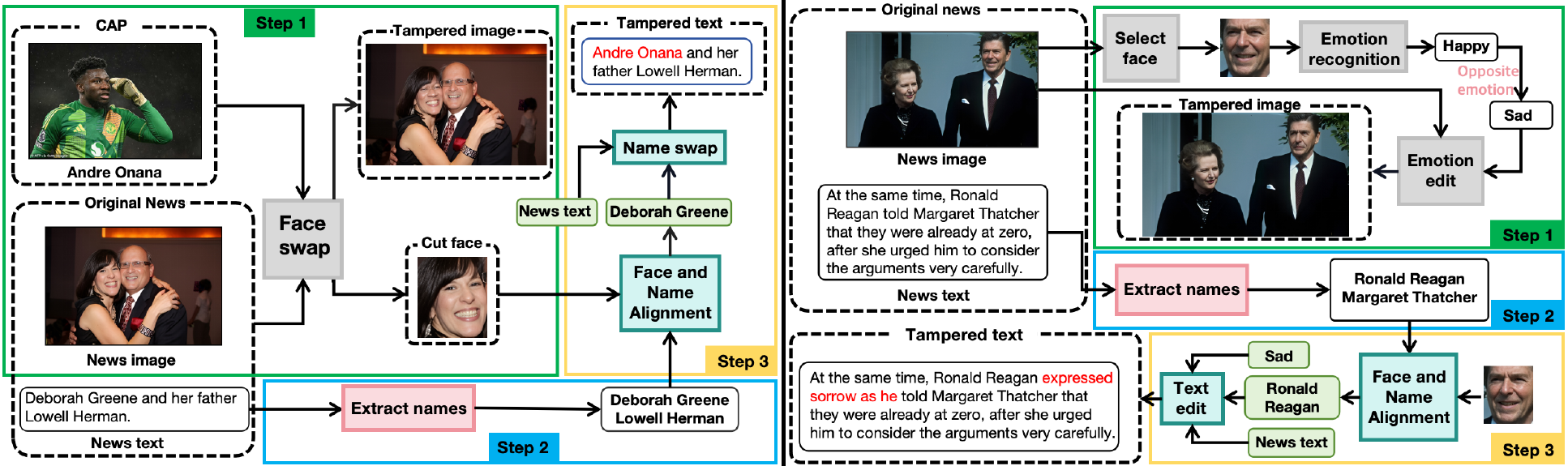}
    \caption{The process of swap manipulation is shown on the left, while the process of attribute manipulation is shown on the right. We use image tampering models and the Qwen series of models to carry out manipulation.}
    \label{fig:dataset_pipeline}
\end{figure*}
\section{SAMM Dataset}
Unlike existing benchmarks that randomly manipulate the multimedia, which result in misaligned semantics in fabricated multimodal media.~\cite{dgm4,newsclipings}. In response, we construct SAMM dataset to remedy this weakness.
The construction of the dataset comprises three steps: 1) Source data collection to filter out data for subsequent tampering operations and build an external knowledge base; 2) Multi-modal manipulation, which details the processes of swap manipulation and attribute manipulation. The construction steps of the dataset are as follows:
%------------------------------------------------------------------------
\subsection{Source Data Collection}
Given the social impact and risks of multimodal news manipulation, we align with DGM$^{4}$~\cite{dgm4} to adopt human-centered news data as raw material. Our dataset is built upon \textbf{VisualNews}~\cite{visualnews} and \textbf{GoodNews}~\cite{goodnews}. To ensure human-centeredness and diversity, the following strategies are adopted to filter the raw data: (1) We select news that at least contains one individual. (2) We encode the news images using CLIP~\cite{clip}, compute their similarity scores with all other images in the dataset, and remove those with high cumulative similarity scores. The filtered dataset $S = \{P_{s} \mid P_{s} = (I_{s}, T_{s})\}$ forms the basis for subsequent dataset construction.

\noindent\textbf{Celeb Attributes Portfolio.} Before diving into the construction of  dataset, we first prepare a Celeb Attributes Protfolio (CAP) to aid the building of SAMM dataset and provide external knowledge. We have collected and curated multimodal data for celebrities from the internet using the Google Search API~\cite{googlesearchapi}, encompassing visual modality information (images) and textual modality information (gender, birth year, occupation, main achievements). CAP covers celebrities featured in datasets such as VisualNews\cite{visualnews}, GoodNews~\cite{goodnews}, DGM$^{4}$~\cite{dgm4}, and SAMM. Furthermore, a celebrity's information card can be acquired by simple name matching.
%--------------------------------------------------------------------

\subsection{Multi-Modal Manipulation}
To ensure the alignment between the manipulated text and image, the image tampering is performed first, and the text fabrication follows. 
We adopt two types of image manipulation: \textbf{Swap Manipulation} and \textbf{Attribute Manipulation}. Specifically, Swap Manipulation includes face replacement in the visual modality and corresponding name replacement in the textual modality, while Attribute Manipulation involves emotion manipulation in the visual modality and corresponding emotion-related vocabulary manipulation in the textual modality. Combined with unaltered original image-text pairs, the dataset comprises three data categories.

\noindent\textbf{Swap Manipulation.} As shown in Figure~\ref{fig:dataset_pipeline}, given $(I,T)$, we employ the existing face-swapping model SimSwap~\cite{simswap} and InfoSwap~\cite{infoswap} for visual manipulation and large language models~\cite{qwen2.5,qwen2-vl} for textual name replacement.

\emph{\textbf{$\triangleright$ Face swap.}} We randomly select a face $I_{f}$ and corresponding name $N_{f}$ from CAP, randomly apply SimSwap~\cite{simswap} or InfoSwap~\cite{infoswap} to $I$ to generate the tampered image $I_{m}$, and record the bounding box coordinates \( (x_1, y_1, x_2, y_2) \) of the swapped face region $F_{m}$ in $I_{m}$. The region correspond to the bbox coordinates in $I$ is denoted as $F$.

\emph{\textbf{$\triangleright$ Text forgery.}} 
To generate text that is semantically aligned with the swapped image, we need to know the name of the face that has been swapped out. To acquire this information,
we utilize the large language model Qwen2.5~\cite{qwen2.5} to extract all names $N = \{N^{i} \mid i = 1,2,\ldots\}$ from $T$ (e.g., ``Joe Biden'', ``Vladimir Putin''). These names and $F$ are fed into the multimodal model Qwen2-VL~\cite{qwen2-vl} to align $F$ with the corresponding name $N^{t}\in N$. We perform manual sampling verification on the name extraction and face matching results generated by the large model to ensure accuracy. (implementation details and accuracy validation are provided in the appendix). Finally, all $N^{t}$ in $T$ are replaced with $N_{f}$ to produce the manipulated text $T_{m}$, accompanied by a one-hot vector ${\mathit{label}}_t$ to indicate whether the $i$-th word is tampered.

\begin{table*}
  \caption{SAMM differentiates from existing deepfake datasets by performing cross-modal and fine-grained manipulation of real news content, providing annotations for manipulated visual regions and tampered textual words.}
  \centering
  \begin{tabular}{@{}lccccccccc@{}}
    \toprule
    \multirow{2}{*}{\textbf{Dataset}} & \multirow{2}{*}{\textbf{Size}} & \multirow{2}{*}{\textbf{Source}} & \multirow{2}{*}{\textbf{Modality}} & \multicolumn{4}{c}{\textbf{Annotations}} & \multirow{2}{*}{\textbf{Cross-modal Mani.}} \\
    \cmidrule(lr){5-8}
    & & & & \textbf{Real/Fake} & \textbf{Mani. Type} & \textbf{BBox} & \textbf{Word Bin.} & \\
    \midrule
    LIAR~\cite{relat14} & 12k+ & Real News & Single & \ding{51} & \ding{55} & \ding{55} & \ding{55} & Unaligned \\
    NewsCLIPpings~\cite{newsclipings} & 980k+ & Real News & Multi & \ding{51} & \ding{55} & \ding{55} & \ding{55} & Unaligned \\
    DeeperForensics-1.0~\cite{deeperforensics} & 60k+ & Social Media & Single & \ding{51} & \ding{55} & \ding{55} & \ding{55} & Unaligned \\
    MiRAGeNews~\cite{miragenews} & 12k+ & Synthesis & Multi & \ding{51} & \ding{55} & \ding{55} & \ding{55} & Unaligned \\
    DGM$^{4}$~\cite{dgm4} & 230k+ & Real News & Multi & \ding{51} & \ding{51} & \ding{51} & \ding{51} & Unaligned \\
    \textbf{SAMM(Ours)} & \textbf{260k+} & \textbf{Real News} & \textbf{Multi} & \textbf{\ding{51}} & \textbf{\ding{51}} & \textbf{\ding{51}} & \textbf{\ding{51}} & \textbf{Aligned} \\
    \bottomrule
  \end{tabular}
  \label{tab:difference}
\end{table*}
\begin{figure*}[t]
    \centering
    \includegraphics[width=\textwidth]{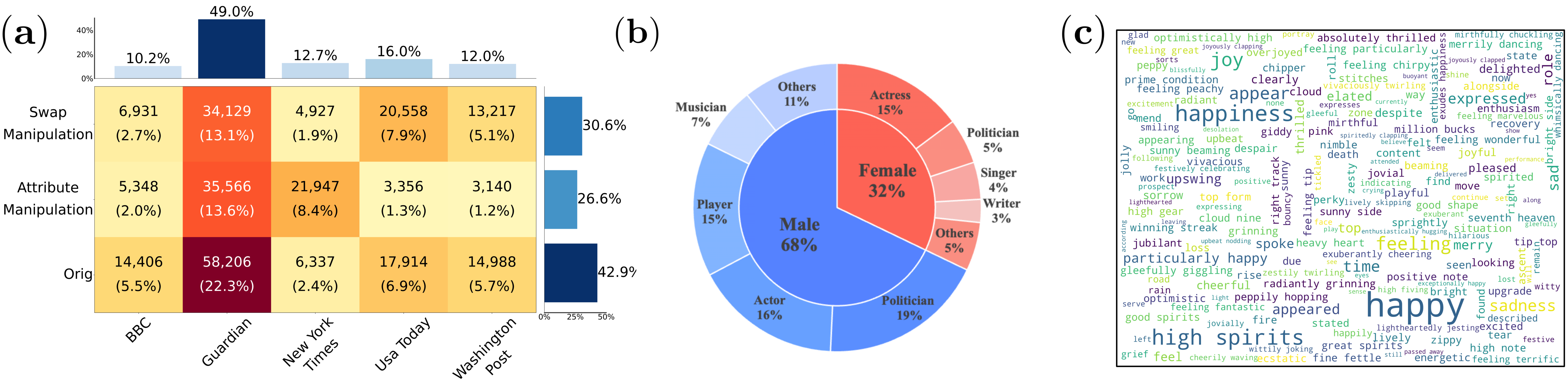}
    \caption{Statistics of the SAMM. (a) The distribution of manipulation types and the distribution of source data; (b) The distribution of gender and occupations among celebrities involved in swap manipulation; (c) The word cloud of emotional descriptions for attribute manipulation.}
    \label{fig:dataset st}
\end{figure*}

\noindent\textbf{Attribute Manipulation.} As shown in figure~\ref{fig:dataset_pipeline}, we employ the HFGI~\cite{hfgi} and StyleCLIP~\cite{styleclip} for visual emotion editing and large language models~\cite{qwen2.5,qwen2-vl} for textual emotion manipulation.

\emph{\textbf{$\triangleright$ Face attribute edit.}} We first utilize the DSFD~\cite{dsfd} model for face detection to randomly select a target face \( F \) in $I$, recording its bounding box coordinates \( (x_1, y_1, x_2, y_2) \). The Qwen2-VL model~\cite{qwen2-vl} predicts the emotional state of \( F \), which is then randomly fed into HFGI~\cite{hfgi} or StyleCLIP~\cite{styleclip} to generate an opposite emotional manipulation (e.g., modifying "happy" to "sad"), resulting in \( F_{m} \).We then replace \( F \) in \( I \) with \( F_{m} \), yielding \( I_m\).

\emph{\textbf{$\triangleright$ Text distortion.}} Following the same operation as in the swap manipulation, we utilize Qwen2.5~\cite{qwen2.5} and Qwen2-vl~\cite{qwen2-vl} to complete the matching of the name \( N^{t} \) and the face \( F \). To achieve diverse emotional expressions, we collect multiple emotions  and provide various expressions for each emotion option $\mathcal{E}=\{e_i|e_i=[w^i_1,w^i_2,...]\}_{i=1}^{|\mathcal{E}|}$, where $w_1^i$ is the 1-th expression word for emotion $e_i$. Subsequently, 
we utilize Qwen2.5~\cite{qwen2.5} to incorporate the randomly selected expression of opposite emotion predicted by the Qwen2-VL~\cite{qwen2-vl} into \( T \) without altering the event described in the news, thereby obtaining \( T_m \), accompanied by a one-hot vector ${\mathit{label}}_t$ to indicate whether the $i$-th word is tampered. To ensure accuracy, we conducted manual sampling inspections on the tasks completed by Qwen2.5 and Qwen2-VL. (Implementation details of Qwen2.5 and Qwen2-VL for these tasks, along with accuracy validation, are provided in the appendix.)
%--------------------------------------------------------------------
\subsection{Dataset Statistics}
The SAMM dataset (260,970 samples) captures real-world tampering patterns with comprehensive annotations. As shown in Figure~\ref{fig:dataset st}(a) and Table~\ref{tab:difference}, it includes: 111K original news, 80K swap-manipulated, and 69K attribute-manipulated cases. Figure~\ref{fig:dataset st}(b) highlights celebrity diversity in swap manipulation, while Figure~\ref{fig:dataset st}(c)'s emotion word clouds demonstrate expressional diversity in attribute manipulation.

%---------------------------------------------------

\section{Methodology}
\begin{figure*}[t]
    \centering
    \includegraphics[width=\textwidth]{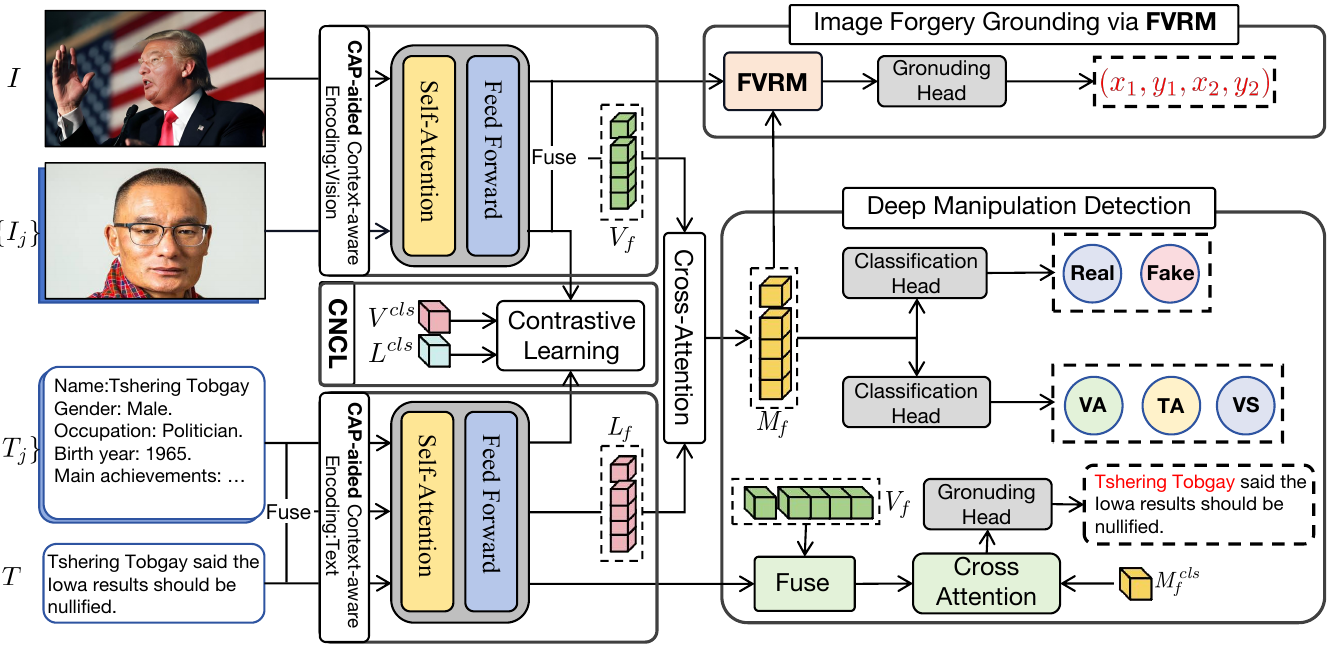}
    \caption{The architecture of our proposed RamDG. It processes image-text pairs and CAP-retrieved knowledge using uni-Encoders and the CNCL module, then fuses knowledge-enriched embeddings for visual/textual manipulation localization, fake news recognition, and manipulation type prediction.}
    \label{fig:ramdg}
\end{figure*}

Figure~\ref{fig:ramdg} depicts the framework of RamDG.
%RamDG includes the information of CAP dataset as auxiliary clues to assist the manipulation detection and grounding. 
Specifically, given multimodal news, the headshot and metadata of the person mentioned in the text are first retrieved from CAP. Next, the multimodal input and retrieved auxiliary input(s) first pass through the CAP-aided Context-aware Encoding, which generates uni-modal embeddings. Subsequently, these embeddings are fed into the Cross-modal Feature Fusion to achieve information fusion across multiple modalities. Image Forgery Grounding, Text Manipulation Localization, Fake News Recognition, and Manipulation Type Recognition are then performed respectively to achieve manipulation detection and grounding. Finally, the overall network is optimized by a combination of grounding and detection losses.

\noindent\textbf{Celebrity Attribute Retrieval.} Given the image-text pair \( P = (I, T) \), we employ a string matching algorithm to rapidly retrieve associated external knowledge from CAP using the person names in \( T \), obtaining a set of related pairs \(\{P_j = (I_j, T_j) | j = 1, 2, \dots, P_j \in \mathit{CAP}\}\) as the auxiliary knowledge.

%------------------------------------------------------------------------
\subsection{CAP-aided Context-aware Encoding}
\textbf{Image Fusion with External Celebrities.} To effectively incorporate the retrieved person images, we first patchify \( I \) and \( \{I_j\} \), then input them into a Transformer-based Vision Encoder \( E_v \)~\cite{vit}, obtaining \( E_v(I) = \mathit{V} = \{\mathit{V}^{cls}, \mathit{V}^{pat}\} \) and \( E_v(I_j) = \mathit{V}_j = \{\mathit{V}_j^{cls}, \mathit{V}_j^{pat}\} \), where \( \mathit{V}^{pat} = \{\mathit{V}^{1}, \mathit{V}^{2},\cdots\} \),\( \mathit{V}_j^{pat} = \{\mathit{V}_j^{1}, \mathit{V}_j^{2},\cdots\} \) are the corresponding patch embeddings. 
To endow the raw multimodal inputs with the knowledge of the focused celebrities, we fuse the  features within modalities to enhance the representative. In particular, for image input $I$,
%We propose \textbf{Visual Knowledge Fusion Mechanism} to better extract and fuse visual information. 
%This fusion mechanism 
we first concatenates the patches \(\{\mathit{V}_j\}\) of all retrieved celebrities to obtain \( \mathit{V}_{cb} \). Subsequently, we query features from $V_{cb}$ using image feature $V$ and then perform a cross-attention~\cite{transformer} to equip the celebrities knowledge to the raw image feature:
%then fuses \( \mathit{V} \) and \(\mathit{V}_{all}\) through a cross-attention mechanism to obtain \( \mathit{V}_{f} \) containing external knowledge. The process is as follows:
%\begin{equation}
%\mathit{V}_{all} = \text{concat}(\mathit{V}_1, \mathit{V}_{2},\cdots,dim=1)
%\end{equation}
\begin{equation}
\begin{split}
\mathit{V}_{f} &= \text{Attn}(Q = \mathit{V}, K = \mathit{V}_{cb}, V = \mathit{V}_{cb}|\Theta_1) %\\
%&= \{\mathit{V}_{f}^{cls}, \mathit{V}_{f}^{pat}\}
\end{split}
\end{equation}
where $\mathit{V}_{f}=\{\mathit{V}_{f}^{cls}, \mathit{V}_{f}^{pat}\}$ is the resultant comprehensive features. Attn($\cdot$) is the multi-head attention function and $\Theta_1$ represents the parameters in the Attn($\cdot$).

\noindent\textbf{Text Encoding with Celebrity Notes.} For text input, as \( \{T_j\} \) have high information density and no redundancy, we directly append the celebrity meta text(s) to the text, thereby directly enriching the textual context. Subsequently, the enhanced text is fed into text encoder $E_t$, thus the compresive representative for text is obtained, marked as $\mathit{L}_{f} = E_t(T_{all}) = \{\mathit{L}_{f}^{cls}, \mathit{L}_{f}^{tok}\}$. The pure text feature $L=\{L^{cls},L^{tok}\}$ and celebrity textual feature $L_f=\{L_f^{cls},L_f^{tok}\}$ are also obtained by feeding the text and the concatenated celebrity notes to text encoder.

\subsection{{Celebrity-News Contrastive Learning}.} 
Human usually doubts a piece of news when they found conflicts between the news and the information they know.
Motivated by this consideration, Celebrity-News Contrastive Learning (CNCL) conducts a contrastive learning~\cite{moco} procedure between the multimodal news and auxiliary celebrity information, aiming to endow the network a human-like reasoning ability for fake news detection.
In specific, we adopt a contrastive learning mechanism to simulate human logical reasoning : by aligning the semantics of untampered celebrity information with the news, we enhance the model's detection ability:
%Below, we first introduce the contrastive learning between external knowledge images and news images. We use InfoNCE Loss, with the specific loss function as follows:
\begin{equation}
\mathcal{L}_{v2v}(I_{j}, I, \mathcal{I}) = -  \log \bigg( 
\frac{\exp(s(I_{j}, I^{+})/\tau)}{\sum_{I_{k} \in \mathcal{I}} \exp(s(I_{j}, I_{k})/\tau)} \bigg) 
\end{equation}
where \( s(\cdot) \) is the cosine similarity function, \(\tau\) is the learnable parameter controlling the temperature, \( I_j \) is the image of the j-th celebrity appearing in the news \( (I, T) \), \( I \) is the corresponding positive sample news image, and $ \mathcal{I} $ is the set of a positive sample news image $I$ and multiple negative samples that do not include the celebrity represented by \( I_j\). We map the \texttt{[CLS]} token through a projection layer  and incorporate it into the cosine similarity calculation as follows:
\begin{equation}
s(I_{j}, I) = [P_{v}(\mathit{V}_{j}^{\mathit{cls}})]^{T}\hat{P}_{v}(\hat{V}^{\mathit{cls}})
\end{equation}
Where \( \hat{\mathit{V}}^{cls} \) is the \texttt{[CLS]} token obtained by encoding \( I \) with  momentum encoder \( \hat{E}_{v} \)~\cite{moco}, $P_{v}$ and $\hat{P}_{v}$ are the mapping layers. Similarly, the contrastive learning $\mathcal{L}_{v2t}(I_j,T,\mathcal{T})$ between external knowledge images and news text can be performed.
In a analogous fashion, we further augment the contrastive learning on the text side, introducing \( \mathcal{L}_{t2v}(T_{j}, I, \mathcal{I}) \) and \(\mathcal{L}_{t2t}(T_{j}, T, \mathcal{T})\). (The specific expressions for $\mathcal{L}_{v2t}, \mathcal{L}_{t2v}, \mathcal{L}_{t2t}$ can be found in the appendix).

\indent In summary, the overall loss function of Celebrity-News Contrastive Learning mechanism is:
\begin{equation}
\mathcal{L}_{cncl} = \mathcal{L}_{v2v} + \mathcal{L}_{v2t} + \mathcal{L}_{t2v} + \mathcal{L}_{t2t}
\end{equation}

%\textbf{Cross-modal Feature Fusion.} Celebrity-News Contrastive Learning mechanism allows the model to examine news image-text pairs like people. Meanwhile, $\mathit{V}_\mathit{f}$ and $\mathit{L}_\mathit{f}$ respectively fuse visual and textual information from external knowledge bases while retaining the semantic information of the original news's visual and textual modalities. 

\subsection{Image Forgery Grounding via FVRM}
To integrate the knowledge of both modalities, we first fuse the information from two modalities to obtain a hybrid multimodal representation with comprehensive contexts, where the text $L_f$ serves as the query to collect clues from image via attention:
\begin{equation}
M_{\mathit{f}} = \text{Attn}(Q = \mathit{V}_{f}, K = \mathit{L}_{f},V = \mathit{L}_{f}|\Theta_2)
\label{eq:Ffus1}
\end{equation}
where $M_{\mathit{f}}=\{\mathit{M}_{\mathit{f}}^{\mathit{cls}}, \mathit{M}_{\mathit{f}}^{\mathit{tok}}\}$.
Since visual manipulation is small-scale and localized, we need to extract local semantic information related to visual tampering. To achieve this, we adopt the Fine-grained Visual Refinement Mechanism (FVRM).

\noindent\textbf{Fine-grained Visual Refinement Mechanism}. Patches in $V^{pat}$ from tampered regions differ semantically from those in unaltered regions, revealing local tampering traces.  Based on this observation, we add a classification head (three linear layers) after $V^{\text{pat}}$ to predict patch manipulation. The loss $\mathcal{L}_{\text{pat}}$ is computed using cross-entropy. Specifically, the loss function is defined as:
\begin{equation}
\mathcal{L}_{pat} = -\sum_{i=1}^{C} \left[ y_{pat} \log(P_{pat}) \right]
\label{eq:Lp_expanded}
\end{equation}
where $y_{pat}$ is the label converted from bbox coordinates to indicate whether a patch is manipulated. $P_{pat}$ represents the probability, predicted by the model, that the patch has been tampered with. $C$ is the number of patches in an image. Under the supervision of $y_{pat}$, $\mathit{V}^{pat}$ is mapped through the first two linear layers of classification head to obtain $\widetilde{V}^{pat}$, which contains visual manipulation traces. $\mathit{M}_{\mathit{f}}^{cls}$ contains semantic information related to the global examination of the detected news pair. By fusing $\widetilde{V}^{pat}$ and $\mathit{M}_{\mathit{f}}^{cls}$, the resulting $\widetilde{M}_{\mathit{f}}$ captures both global and local information. The process is as follows:
\begin{equation}
\widetilde{M}_{\mathit{f}} = \text{Attn}(Q = \mathit{M}_{\mathit{f}}^{\text{cls}}, \, K = \widetilde{V}^{\text{pat}}, \, V = \widetilde{V}^{\text{pat}}|\Theta_3)
\label{eq:Ffus2}
\end{equation}
%To adapt to the extraction of fine-grained visual tampering semantic information, we further introduce a multi-scale information representation mechanism: 
Let $\widetilde{M}_{\mathit{f}}$ perform attention computation with a learnable vector $\mathcal{Q}$~\cite{blip2} to capture semantic information at different scales. The fusion of external knowledge in $M_{\mathit{f}}$ dilutes the local details of the original image. To address this, we perform residual connection~\cite{resnet}, as follows:
\begin{equation}
\mathit{M}_{\mathit{fv}} = M_{\mathit{f}} + \mathit{V}
\label{eq:Vgru}
\end{equation}

Finally, cross-attention is performed between \(\widetilde{M}_{\mathit{f}}\) and \(\mathit{M}_{\mathit{fv}}\) to achieve fine-grained visual refinement, denoted as \(\widetilde{M}_{\mathit{fv}}\). The specific process is as follows:
\begin{equation}
\widetilde{M}_{\mathit{f}} = \text{Attn}( Q = \mathcal{Q}, \,
K = \widetilde{M}_{\mathit{f}}, V = \widetilde{M}_{\mathit{f}}|\Theta_4)
\label{eq:Ffus3}
\end{equation}
\begin{equation}
\widetilde{M}_{\mathit{fv}} = \text{Attn}(Q = \widetilde{M}_{\mathit{f}}, 
K = {M}_{\mathit{fv}}, V = {M}_{\mathit{fv}}|\Theta_5)
\label{eq:Ffus4}
\end{equation}

We use the obtained $ \widetilde{M}_{\mathit{fv}} $ for bounding box prediction. The $L_1$ loss and IoU loss~\cite{iou} between the predicted boxes and the bounding box coordinates are then computed:
\begin{equation}
\begin{aligned}
\mathcal{L}_\mathit{bbox} = -||P_{gro}-\text{Sigmoid}(y_{box})||_1 \\
+ \mathcal{L}_\mathit{IoU}(P_{gro}, \text{Sigmoid}(\mathit{y}_{box}))
\end{aligned}
\label{eq:Lgr}
\end{equation}
where $\mathcal{L}_\mathit{IoU}(\cdot)$ is the IoU loss function, $P_{gro}$ represents the predicted bounding box coordinates from the model.
\subsection{Deep Manipulation Detection}
\noindent\textbf{Text Manipulation Localization.} This section achieves fine-grained text manipulation localization: predicting whether each word in $T$ is replaced or added. 
%$\mathit{L}^{\mathit{tok}}$ relies solely on internal textual context, lacking cross-modal logical verification. To address this, 
We first fuse the textual representation $\mathit{L}^{\mathit{tok}}$ with the visual semantic feature $\mathit{V}_{\mathit{f}}$ through cross-attention to generate a fine-grained cross-modal feature $\mathit{L}_{\mathit{v}}^{\mathit{tok}}$:
\begin{equation}
\mathit{L}_{\mathit{v}}^{\mathit{tok}} = \text{Attn}(Q = \mathit{L}^{\mathit{tok}}, 
K = \mathit{V}_{\mathit{f}}, V = \mathit{V}_{\mathit{f}}|\Theta_5)
\label{eq:Ffus5}
\end{equation}

$M_{\mathit{f}}^{\mathit{cls}}$ can locate tokens in $\mathit{L}_{\mathit{v}}^{\mathit{tok}}$ with abnormal logical relationships due to its global semantic information, which are often added or modified. Therefore, we further allow $M_{\mathit{f}}^{\mathit{cls}}$ to extract information from $\mathit{L}_{\mathit{v}}^{\mathit{tok}}$. The above process is as follows:
\begin{equation}
\widetilde{L}_{\mathit{v}}^{\mathit{tok}} = \text{Attn}(Q = \mathit{L}_{\mathit{v}}^{\mathit{tok}},
K = \mathit{M}_{\mathit{f}}^{\mathit{cls}}, V = \mathit{M}_{\mathit{f}}^{\mathit{cls}}|\Theta_6)
\label{eq:Ffus6}
\end{equation}

Finally, we use $\widetilde{L}_{\mathit{v}}^{\mathit{tok}}$ to predict the probability of manipulation for each token, and construct the text manipulation localization loss function $\mathcal{L}_\mathit{tok}$ based on the cross-entropy loss function:
\begin{equation}
\mathcal{L}_\mathit{tok} = -\sum_{i} \left[ y_{tok}^i \log(P_{tok}^i) \right]
\label{eq:Ltx}
\end{equation}
where $y_{\mathit{tok}}^i$ is the label indicating whether i-th token is manipulated. $P_{tok}^i$ represents the probability, as output by the model, that the i-th token has been tampered.

\noindent\textbf{Fake News Recognition.} This section is used to determine whether $(I, T)$ has been tampered. We utilize $M_f^{\mathit{cls}}$ for recognition, as it integrates multi-modal news data and external knowledge from CAP to provide a comprehensive summary. We compute the loss using a cross-entropy function, as detailed below:
\begin{equation}
\mathcal{L}_{2cls} =  -y_{2cls}\log(P_{2cls}) 
\label{eq:L2cls}
\end{equation}
where $y_{2cls}$ is the binary classification label and $P_{2cls}$ represents the probability, predicted by the model, that $(I, T)$ has been tampered with.

\noindent\textbf{Maniputation Type Recognition.} SAMM incorporates three types of tampering traces, including \textbf{V}isual \textbf{S}wap(VS) manipulation, \textbf{V}isual \textbf{A}ttribute(VA) manipulation, \textbf{T}extual \textbf{A}ttribute(TA) manipulation (including name swapping, and the addition or alteration of emotions). We predict the specific tampering method used in the given news.  Consistent with Fake News Recognition, we uses $M_f^{\mathit{cls}}$ to predict and compute the loss using a cross-entropy function, as detailed below:
\begin{equation}
\mathcal{L}_{\mathit{mcls}} = -\sum_{i=1}^4 \left[ y_{mcls}^i \log(P_{mcls}^i) \right]
\label{eq:Lcls}
\end{equation}
where $\mathit{y}_{mcls}^i$ denotes the label of the i-th tampering method used in the given news. $P_{mcls}^i$ represents the probability, predicted by the model, that the i-th tampering method has been used.
%------------------------------------------------------------------------
\begin{table*}[ht]
\caption{Comparison of multi-modal learning methods for SAMM. We conduct a comprehensive evaluation of models' performance across four key tasks: binary classification, multi-label classification, image grounding, and text grounding.}
\begin{tabular}{l|ccc|ccc|ccc|ccc}
\toprule
\multicolumn{1}{c}{} & \multicolumn{3}{c}{Binary Cls(BC)} & \multicolumn{3}{c}{Multi-Label Cls(MLC)} & \multicolumn{3}{c}{Image Grounding(IG)} & \multicolumn{3}{c}{Text Grounding(TG)} \\
\cmidrule(lr){2-4} \cmidrule(lr){5-7} \cmidrule(lr){8-10} \cmidrule(lr){11-13}
\multicolumn{1}{l}{\multirow{2}{*}[14pt]{\textbf{Methods}}} & \multicolumn{1}{c}{AUC} & \multicolumn{1}{c}{EER} & \multicolumn{1}{c}{ACC} & \multicolumn{1}{c}{mAP} & \multicolumn{1}{c}{CF1} & \multicolumn{1}{c}{OF1} & \multicolumn{1}{c}{IoUmean} & \multicolumn{1}{c}{IoU50} & \multicolumn{1}{c}{IoU75} & \multicolumn{1}{c}{Precision} & \multicolumn{1}{c}{Recall} & \multicolumn{1}{c}{F1} \\
\midrule
VILT~\cite{vilt} & 96.10 & 11.02 & 88.83 & 96.03 & 90.21 & 89.84 & 65.38 & 71.91 & 54.49 & 77.42 & 69.78 & 73.40 \\
HAMMER~\cite{dgm4} & 97.85 & 7.80 & 92.43 & 97.98 & 93.77 & 93.44 & 77.68 & 84.41 & 78.44 & 85.94 & 82.74 & 84.31 \\
HAMMER++~\cite{hammer++} & 97.60 & 7.99 & 92.26 & 97.72 & 93.70 & 93.34 & 77.66 & 84.12 & 78.62 & 85.86 & 82.89 & 84.35 \\
FKA-Owl~\cite{fka-owl} & 98.09 & 7.19 & 92.60 & 2.53 & 13.97 & 13.84 & 66.40 & 73.54 & 54.82 & 19.16 & 49.71 & 27.66 \\
Qwen2.5VL-72b~\cite{qwen2.5-vl} & 76.67 & 44.93 & 55.06 & \textemdash & \textemdash & \textemdash & \textemdash & \textemdash & \textemdash & \textemdash & \textemdash & \textemdash \\
\textbf{RamDG(Ours)} & \textbf{98.79} & \textbf{5.42} & \textbf{94.66} & \textbf{98.86} & \textbf{95.52} & \textbf{95.33} & \textbf{80.90} & \textbf{87.56} & \textbf{82.00} & \textbf{86.16} & \textbf{83.54} & \textbf{84.83} \\
\bottomrule
\end{tabular}
\label{tab:performance}
\end{table*}
%-------------------------------------------------------
\begin{table}[ht]
\caption{Comparison on the SAMM 20000 training set.}
\begin{tabular}{l|cccc}
\toprule
\multicolumn{1}{l}{\multirow{2}{*}[-2pt]{\textbf{Methods}}} 
& \multicolumn{1}{c}{BC} & \multicolumn{1}{c}{MLC} & \multicolumn{1}{c}{IG} & \multicolumn{1}{c}{TG} \\
\cmidrule{2-5} 
\multicolumn{1}{l}{} 
& \multicolumn{1}{c}{ACC} & \multicolumn{1}{c}{mAP} & \multicolumn{1}{c}{IoUmean} & \multicolumn{1}{c}{F1} \\
\midrule
VILT~\cite{vilt} & 81.97 & 87.31 & 39.95 & 51.15 \\
HAMMER~\cite{dgm4} & 85.74 & 92.98 & 52.65 & 71.65 \\
HAMMER++~\cite{hammer++} & 84.03 & 92.72 & 58.61 & 72.10 \\
FKA-Owl~\cite{fka-owl} & 87.91 & 1.94 & 49.48 & 9.79 \\
Qwen2.5VL-72b~\cite{qwen2.5-vl} & 55.06 & \textemdash & \textemdash & \textemdash \\
\textbf{RamDG(Ours)} & \textbf{88.40} & \textbf{95.32} & \textbf{64.30} & \textbf{73.69} \\
\bottomrule
\end{tabular}
\label{tab:performance 2w}
\end{table}

\begin{table}[ht]
\caption{Additional VLMs' performence on the SAMM.}
\centering
\label{tab:vlms_performance}
\begin{tabular}{lcc}
\toprule
\multicolumn{1}{c}{\textbf{Methods (BC/ACC)}} & \multicolumn{1}{c}{Zero-Shot} & \multicolumn{1}{c}{Finetuned} \\
\midrule
LLaMA-3.2-Vision-90B~\cite{llama3.2}   & 60.4 & -- \\
Gemini-3-27B~\cite{Gemma3}       & 59.7 & -- \\
SeedVL-1.5~\cite{seedvl1.5}          & \textbf{64.1} & -- \\
\cmidrule(lr){1-3}
Qwen2.5VL-3B~\cite{qwen2.5-vl}         & -- & 82.0 \\
\textbf{RamDG(Ours)}                & -- & \textbf{94.66} \\
\bottomrule
\end{tabular}
\end{table}

\subsection{Overall Loss Function}
The overall loss function for the training process is as follows:
\begin{equation}
\mathcal{L} = \mathcal{L}_{\mathit{cncl}} +  \mathcal{L}_{2\mathit{cls}} + \mathcal{L}_{\mathit{mcls}} + \mathcal{L}_{\mathit{pat}} + \mathcal{L}_{\mathit{bbox}} + \mathcal{L}_{\mathit{tok}}
\label{eq:total_loss}
\end{equation}
%--------------------------------------------------------------
\section{Experiments}
%-----------------------------------------------------------------
Please refer to the supplementary material for implementation details and evaluation metrics.

\noindent\textbf{Comparison Methods}. We selected three modailty fusion-based methods — VILT~\cite{vilt}, HAMMER~\cite{dgm4} and HAMMER++~\cite{hammer++}, along with the state-of-the-art Visual-Language Large Models(VLLMs) — FKA-Owl~\cite{fka-owl} and Qwen2.5VL-72b~\cite{qwen2.5-vl} as baselines for performance comparison with RamDG on the SAMM. Implementation details for the methods can be found in the Appendix. 

\subsection{Quantitative Results}

\textbf{Performance Comparison}. Table~\ref{tab:performance} shows the performance of all the aforementioned baselines on the SAMM dataset. To simulate real-world scenarios with scarce training samples, we train these models on randomly selected subsets of 20,000 and 50,000 samples. We then evaluate their performance on the complete test set, as shown in Table~\ref{tab:performance 2w} and Table~\ref{tab:performance 5w}. Experimental results showed in tables prove that our method  achieved state-of-the-art performance across various tasks on the SAMM dataset. Notably, under conditions of limited training data, RamDG demonstrated significant advantages over baseline models, particularly in the precision of visual tampering region localization, fully demonstrating the effectiveness and superiority of our proposed method.

While FKA-Owl slightly outperforms HAMMER in binary classification (+0.24\%), it fails at fine-grained tampering localization. Though VLLMs leverage rich knowledge for fake news judgment, they lack fine-grained extraction capability. In contrast, our RamDG: 1) retrieves CAP knowledge, 2) integrates it via CNCL, and 3) enhances visual localization with FVRM – achieving superior performance across all tasks.

\textbf{More VLMs' performance}. We conduct evaluations on additional VLMs under both zero-shot and fine-tuned settings, as shown in the table~\ref{tab:vlms_performance}.

\textbf{Generalization to Unseen Entities}. To evaluate new entities absent from CAP, we select a sub-test set whose entities absent from the training set and directly input them into RamDG without retrieving information from CAP. As shown in the table~\ref{tab:generalization to unseen entities}, our RamDG still outperforms comparison methods.
%-------------------------------------------------------
\begin{table}[t]
\caption{Comparison on the SAMM 50000 training set.}
\begin{tabular}{l|cccc}
\toprule
\multicolumn{1}{l}{\multirow{2}{*}[-2pt]{\textbf{Methods}}} 
& \multicolumn{1}{c}{BC} & \multicolumn{1}{c}{MLC} & \multicolumn{1}{c}{IG} & \multicolumn{1}{c}{TG} \\
\cmidrule{2-5} 
\multicolumn{1}{l}{} 
& \multicolumn{1}{c}{ACC} & \multicolumn{1}{c}{mAP} & \multicolumn{1}{c}{IoUmean} & \multicolumn{1}{c}{F1} \\
\midrule
VILT~\cite{vilt} & 85.18 & 92.73 & 55.50 & 64.91 \\
HAMMER~\cite{dgm4} & 88.16 & 95.58 & 65.97 & 76.02 \\
HAMMER++~\cite{hammer++} & 87.99 & 95.29 & 68.15 & 78.48 \\
FKA-Owl~\cite{fka-owl} & 90.36 & 1.19 & 63.28 & 27.66 \\
Qwen2.5VL-72b~\cite{qwen2.5-vl} & 55.06 & \textemdash & \textemdash & \textemdash \\
\textbf{RamDG(Ours)} & \textbf{91.07} & \textbf{97.18} & \textbf{73.65} & \textbf{79.10} \\
\bottomrule
\end{tabular}
\label{tab:performance 5w}
\end{table}

\begin{table}[t]
\caption{Generalization to unseen entities.}
\centering
\label{tab:generalization to unseen entities}
\begin{tabular}{lcccc}
\toprule
\multicolumn{1}{c}{\textbf{Methods}} & \multicolumn{1}{c}{ACC} & \multicolumn{1}{c}{mAP} & \multicolumn{1}{c}{IoUmean} & \multicolumn{1}{c}{F1} \\
\midrule
HAMMER~\cite{dgm4}    & 92.0 & 97.0 & 77.6 & 83.8 \\
FKA-Owl~\cite{fka-owl}   & 92.3 & 4.0  & 68.1 & 28.1 \\
\textbf{RamDG(Ours)}     & \textbf{94.1} & \textbf{97.3} & \textbf{78.7} & \textbf{83.9} \\
\bottomrule
\end{tabular}
\end{table}

%----------------------------------------------------
\begin{table*}[t]
\caption{Ablation study for external knowledge from CAP. For each task, we present the most representative metrics: ACC, mAP, IoUmean, and F1.}
\centering
\begin{tabular}{ccccc|cccc}
\toprule
\multicolumn{5}{c}{External Knowledge From CAP} & \multicolumn{1}{c}{BC} & \multicolumn{1}{c}{MLC} & \multicolumn{1}{c}{IG} & \multicolumn{1}{c}{TG} \\
\cmidrule(lr){1-5} \cmidrule(lr){6-9}
\multicolumn{1}{c}{Gender} & \multicolumn{1}{c}{Birth Year} & \multicolumn{1}{c}{Occupation} & \multicolumn{1}{c}{Main Achievements} & \multicolumn{1}{c}{Images} & \multicolumn{1}{c}{ACC} & \multicolumn{1}{c}{mAP} & \multicolumn{1}{c}{IoUmean} & \multicolumn{1}{c}{F1} \\
\midrule
 &  &  &  &  & 91.00 & 96.88 & 75.77 & 83.49 \\
 &  &  &  & \ding{51} & 91.17 & 97.01 & 76.79 & 83.26 \\
\ding{51} & \ding{51} & \ding{51} & \ding{51} &  & 93.73 & 98.08 & 79.49 & \textbf{84.84} \\
 & \ding{51} & \ding{51} & \ding{51} & \ding{51} & 94.32 & 98.21 & 79.84 & 84.10  \\
\ding{51} &  & \ding{51} & \ding{51} & \ding{51} & 94.57 & 98.35 & 80.59 & 84.50 \\
\ding{51} & \ding{51} &  & \ding{51} & \ding{51} & 93.25 & 98.15 & 77.96 & 84.12 \\
\ding{51} & \ding{51} & \ding{51} &  & \ding{51} & 93.48 & 98.34 & 79.46 & 84.69 \\
\ding{51} & \ding{51} & \ding{51} & \ding{51} & \ding{51} & \textbf{94.66} & \textbf{98.86} & \textbf{80.90} & 84.83 \\
\bottomrule
\end{tabular}
\label{tab:cap}
\end{table*}
%------------------------------------------------------
\begin{table}[ht]
\caption{Ablation study for CNCL and FVRM.}
\centering
\begin{tabular}{cc|cccc}
\toprule
\multicolumn{2}{c}{Module} & \multicolumn{1}{c}{BC} & \multicolumn{1}{c}{MLC} & \multicolumn{1}{c}{IG} & \multicolumn{1}{c}{TG} \\
\cmidrule(lr){1-2} \cmidrule(lr){3-6}
\multicolumn{1}{c}{CNCL} & \multicolumn{1}{c}{FVRM} & \multicolumn{1}{c}{ACC} & \multicolumn{1}{c}{mAP} & \multicolumn{1}{c}{IoUmean} & \multicolumn{1}{c}{F1} \\
\midrule
 & \ding{51} & 93.24 & 98.18 & 79.32 & 84.23 \\
\ding{51} &  & 94.79 & 98.88 & 78.01 & 85.28 \\
\ding{51} & \ding{51} & 94.66 & 98.86 & \textbf{80.90} & 84.83 \\
\bottomrule
\end{tabular}
\label{tab:performance fvrm}
\end{table}
\subsection{Ablation study}
\textbf{External knowledge from CAP}. To investigate the impact of different celebrity information in CAP on model performance, we conducted a series of ablation experiments. The results, as shown in Table~\ref{tab:cap}, reveal several key observations: 

1) Without leveraging CAP-derived external knowledge, model performance across tasks drops by an average of 3\%. Single-modal external knowledge alone is insufficient: textual knowledge is indispensable, with its absence causing significant declines in fake news detection (4.11\% drop) and even visual localization. In contrast, visual knowledge provides minimal improvement (0.78\% average gain) due to image redundancy.

2) Analysis shows textual knowledge components affect performance variably. Occupation information contributes the most (1.44\% average gain), reflecting its role in providing contextual and social cues for human verification. For example, knowing Messi’s occupation helps debunk false claims like his Nobel Prize win.
 
\noindent\textbf{Framework Component Ablation.} As shown in Table~\ref{tab:performance fvrm}, we investigated the impact of CNCL and FVRM on model performance. After removing the FVRM module, we directly used $M_{fv}^{cls}$ for visual tampering localization. The results reveal several key insights: 

1) Removing CNCL reduces performance across all tasks (avg. -1.07\%), confirming its role in enhancing external knowledge understanding for multimodal tampering detection.

2) FVRM specifically boosts visual tampering localization (+2.89\%) with minimal impact on other tasks, demonstrating its fine-grained visual tampering capture capability.
\subsection{Visualized results}
\begin{figure}[h]
    \centering
    \includegraphics[width=1\linewidth]{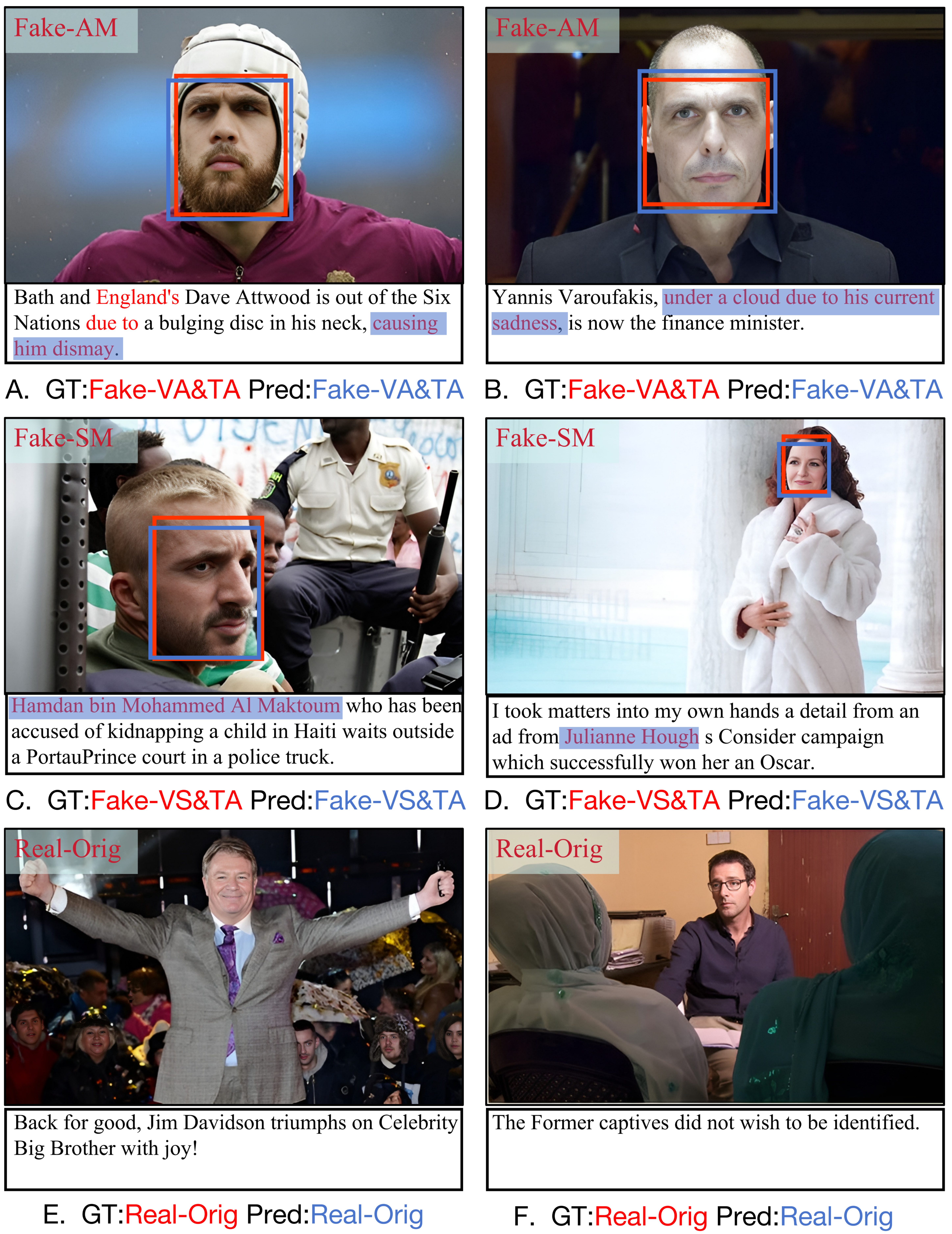}
    \caption{Visualized results. Red and blue regions denote the ground truth and predicted outputs, respectively.}
    \label{fig:visialization}
\end{figure}

Figure~\ref{fig:visialization} presents results for six cases: Examples A–B (Attribute Manipulation, AM) involve Visual Attribute (VA) and Textual Attribute (TA) manipulations; C–D (Swap Manipulation, SM) feature Visual Swap (VS) and TA manipulations; E–F represent original, unmanipulated news.

%The six cases demonstrate that our model can accurately determine the authenticity of a news while providing precise identification of manipulation types and localization of manipulated regions across multiple modalities.

\section{Conclusion}
In this paper, we propose a realistic research scenario: detecting and grounding
semantic-coordinated multimodal manipulations, and introduce a new dataset SAMM. To address this challenge, we design the RamDG framework, proposing a novel approach for detecting fake news by leveraging external knowledge, which consists of two core components: CNCL and FVRM.
%1) CNCL mechanism, which mimics human thinking to fully leverage external knowledge from CAP; 2) FVRM module, dedicated to detecting fine-grained visual tampering traces.
Extensive experimental results demonstrate the effectiveness of our approach.

\begin{acks}
The paper is supported by  the Fundamental Research Funds for the Central Universities with No. JZ2024HGTB0261 and the NSFC project under grant No. 62302140. The computation is completed on the HPC Platform of Hefei University of Technology.
\end{acks}

\bibliographystyle{ACM-Reference-Format}
\balance
\bibliography{sample-base}

\appendix
\clearpage
\section{Details for CAP}
\begin{figure}[h]
    \centering
    \includegraphics[width=1\linewidth]{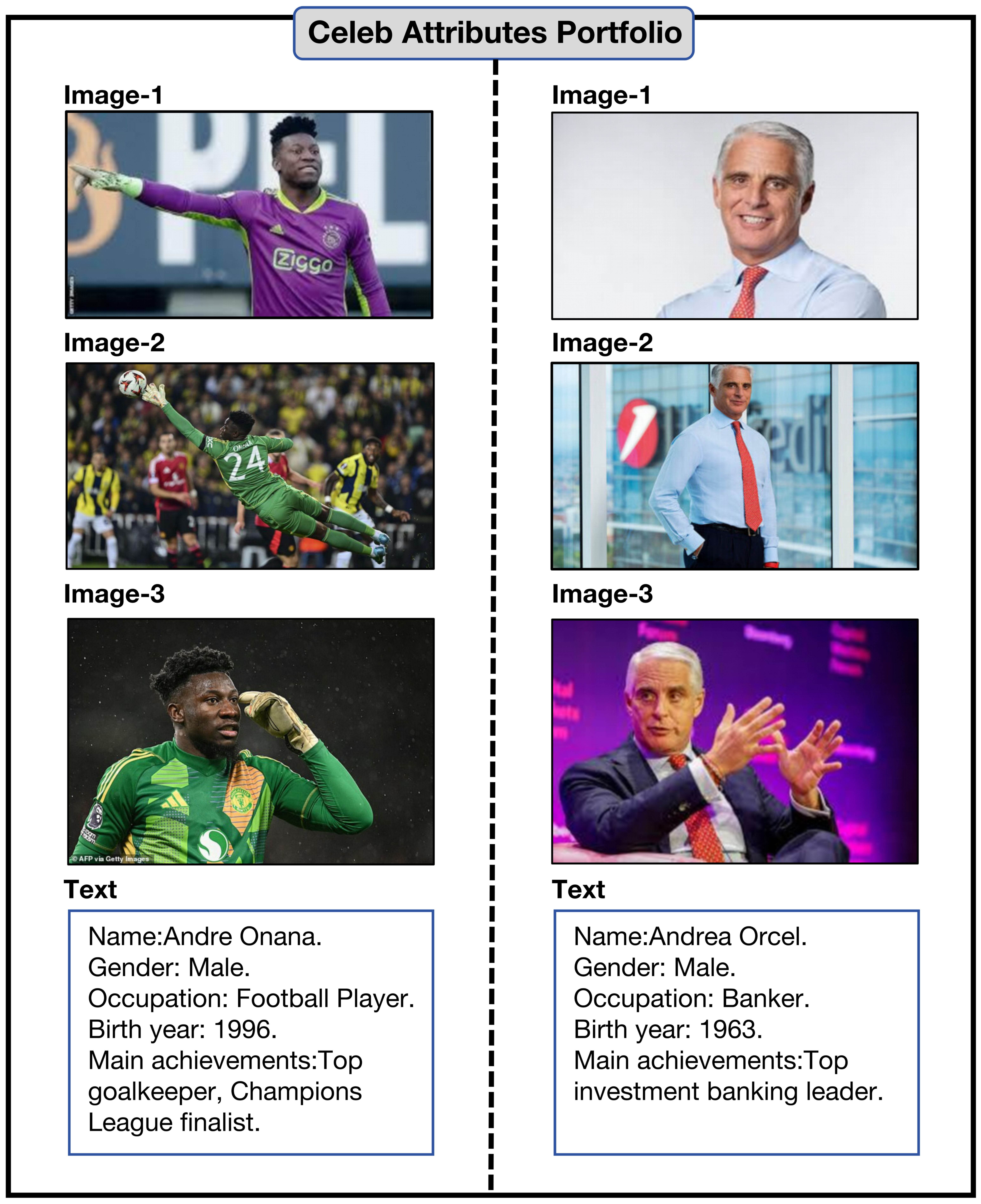}
    \caption{Two examples from CAP}
    \label{fig:cap}
\end{figure}
Each celebrity in the CAP has three associated images along with their gender, birth year, occupation, and main achievements. Figure~\ref{fig:cap} provides two examples from CAP.
\section{Implementation details}
The image encoder is realized using the ViT-B/16 model, which consists of 12 layers. The text encoder are constructed based on a 6-layer transformer. The contrastive learning framework utilizes a queue length of 65,536. During training, we randomly select one image from the three photos of a celebrity in CAP as the visual external knowledge. During testing, we choose the first image from the three photos as the visual external knowledge. Both the classification heads and localization heads are constructed using two-layer MLPs (Multi-Layer Perceptrons). For optimization, the AdamW optimizer is employed with the following parameter settings: the base learning rate for text is set to $5 \times 10^{-6}$, while the learning rate for images is set to $2.5 \times 10^{-5}$. The weight decay is configured at 0.02. The scheduler uses a cosine annealing strategy with an initial learning rate of $5 \times 10^{-6}$ over 50 epochs. The minimum learning rate is set to $2.5 \times 10^{-6}$, and the decay rate is kept at 1. A warm-up phase is included, where the learning rate gradually increases from $2.5 \times 10^{-7}$ over the first 10 epochs. No cooldown period is applied. All experiments were conducted on either 4 NVIDIA GeForce RTX 4090 GPUs or 2 NVIDIA A100 40GB GPUs.
\section{Evaluation metrics}
For binary classification problems, we use the Area Under the Receiver Operating Characteristic Curve (AUC), Accuracy (ACC), and Error Rate (ERR). For multi-class classification problems, we adopt the Mean Average Precision (mAP), Overall F1-score (OF1), and Class-wise F1-score (CF1). For visual tampering localization, we utilize the mean Intersection over Union (IoUmean), Intersection over Union at 50\% threshold (IoU50), and Intersection over Union at 75\% threshold (IoU75). For textual tampering localization, we employ Accuracy (ACC), Recall, and F1-score (F1).
\section{Specific expressions for CNCL}
The specific form of the CNCL loss function is as follows:
\begin{equation}
\mathcal{L}_{v2v}(I_{j}, I, \mathcal{I}) = -  \log \bigg( 
\frac{\exp(s(I_{j}, I^{+})/\tau)}{\sum_{I_{k} \in \mathcal{I}} \exp(s(I_{j}, I_{k})/\tau)} \bigg) 
\end{equation}
\begin{equation}
\hspace{-0.7em}
\mathcal{L}_{v2t}(I_{j}, T, \mathcal{T}) = -  \log \bigg(\frac{\exp(s(I_{j}, T^{+})/\tau)}{\sum_{T_{k} \in \mathcal{T}} \exp(s(I_{j}, T_{k})/\tau)} \bigg) 
\end{equation}
\begin{equation}
\hspace{-0.7em}
\mathcal{L}_{t2v}(T_{j}, I, \mathcal{I}) = -  \log \bigg( 
\frac{\exp(s(T_{j}, I^{+})/\tau)}{\sum_{I_{k} \in \mathcal{I}} \exp(s(T_{j}, I_{k})/\tau)} \bigg) 
\end{equation}
\begin{equation}
\hspace{-0.8em}
\mathcal{L}_{t2t}(T_{j}, T, \mathcal{T}) = -  \log \bigg( 
\frac{\exp(s(T_{j}, T^{+})/\tau)}{\sum_{T_{k} \in \mathcal{T}} \exp(s(T_{j}, T_{k})/\tau)} \bigg) 
\end{equation}
\begin{equation}
\mathcal{L}_{cncl} = \mathcal{L}_{v2v} + \mathcal{L}_{v2t} + \mathcal{L}_{t2v} + \mathcal{L}_{t2t}
\end{equation}
\section{Details for using Qwen2.5 and Qwen2-VL in SAMM construction.}
\begin{figure*}
    \centering
    \includegraphics[width=\textwidth]{qwen2.5samm.pdf}
    \caption{Prompts to guide Qwen2.5 in extracting names and adding emotional descriptions.}
    \label{fig:qwen2.5samm}
\end{figure*}
\begin{figure*}
    \centering
    \includegraphics[width=\textwidth]{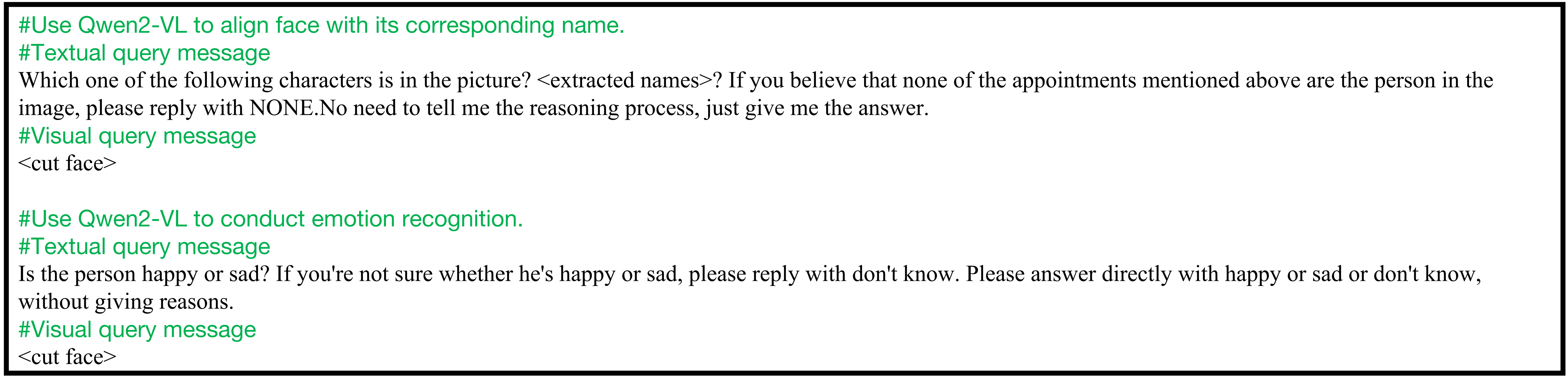}
    \caption{Prompts to guide Qwen2-VL in matching names to faces and performing emotion recognition.}
    \label{fig:qwen2-vlsamm}
\end{figure*}
\textbf{Implementation details}. Figures~\ref{fig:qwen2.5samm} and~\ref{fig:qwen2-vlsamm} show the prompts we used for completing a series of dataset construction tasks with Qwen2.5 and Qwen2-VL, respectively.\\
\textbf{Accuracy validation}. To ensure data quality, we manually inspected the results of each task performed by Qwen2.5 and Qwen2-VL: We randomly selected 500 samples from each task for review, and ultimately, over 95\% of the samples met the required standards in all tasks.

\section{Implementation details for the methods on the SAMM.}
\begin{figure*}
    \centering
    \includegraphics[width=\textwidth]{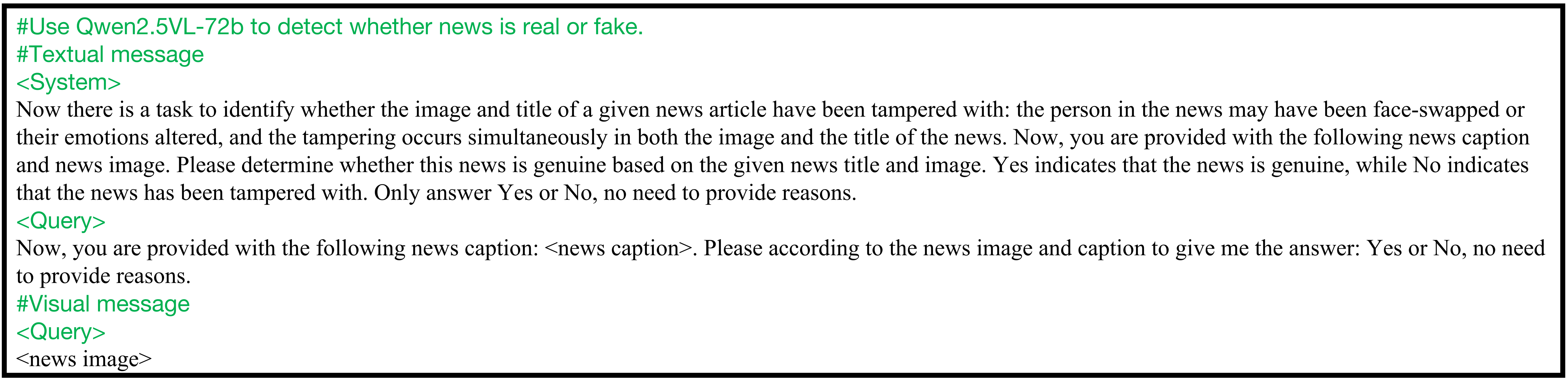}
    \caption{A prompt to guide the testing of Qwen2.5VL-72b on the SAMM dataset.}
    \label{fig:qwen2.5-vl}
\end{figure*}
For VILT, we incorporated detection and localization heads to allow it to identify and localize tampered news content. For FKA-Owl, which has supervision for tampered region coordinates, we added a text tampering localization head after the text embedding stage and appended a manipulation type prediction head following the multimodal feature extraction. For Qwen2.5VL-72b, We use a prompt to guide the model in determining whether a given image-text pair from the test set has been tampered. Figure~\ref{fig:qwen2.5-vl} shows the prompt we used for testing Qwen2.5VL-72b on the SAMM.

\section{Computational Complexity and Efficiency Analysis.}
We test the inference cost on an NVIDIA RTX 4090. The results are reported in the following table~\ref{tab:efficiency_comparison}, where FLOPs and Latency represent per-sample values. Compared with other baselines, our method still achieves an acceptable inference speed.
\begin{table*}[ht]
\caption{Computational efficiency comparison.}
\centering
\label{tab:efficiency_comparison}
\begin{tabular}{lcccc}
\toprule
\textbf{Metric} & \textbf{Params (M)} & \textbf{FLOPs (G)} & \textbf{Latency (ms)} & \textbf{GPU Mem (MB)} \\
\midrule
HAMMER++~\cite{hammer++} & 199.1 & 49.7 & 2.7 & 5874 \\
FKA-Owl~\cite{fka-owl}  & 6771 & 3656.7 & 489.4 & 18772 \\
\textbf{RamDG(Ours)}    & 203.24 & 53.0 & 12.3 & 7357 \\
\bottomrule
\end{tabular}
\end{table*}
\end{document}